# Uniform Learning in a Deep Neural Network via "Oddball" Stochastic Gradient Descent

Andrew J.R. Simpson [#1]

[#] *Centre for Vision, Speech and Signal Processing, University of Surrey*
*Surrey, UK*
[1] `Andrew.Simpson@Surrey.ac.uk`

*Abstract*— **When training deep neural networks, it is typically assumed that the training examples are *uniformly difficult to learn*. Or, to restate, it is assumed that the training error will be uniformly distributed across the training examples. Based on these assumptions, each training example is used an equal number of times. However, this assumption may not be valid in many cases. "Oddball SGD" (novelty-driven stochastic gradient descent) was recently introduced to drive training probabilistically according to the error distribution – training frequency is proportional to training error magnitude. In this article, using a deep neural network to encode a video, we show that *oddball SGD* can be used to enforce uniform error across the training set.**

## I. INTRODUCTION

When training a deep neural network (DNN) with stochastic gradient descent (SGD), some assumptions are made of the training data with respect to the learning that is necessary. In particular, it is assumed that learning should be performed uniformly across the training set. This means that all the examples of the training set receive an equal number of steps to update the weights during training. However, in practice, there is no reason that this assumption of uniformity should hold for DNN whose learning is arbitrarily non-convex and whose training data are arbitrarily distributed.

A recent approach to this problem [1, see 2] is to drive the path of SGD according to the evolving error distribution across the training set. Each training example is assigned a selection probability that is proportional to the error magnitude. The result of this is a *novelty-driven* SGD known as "oddball SGD" [1]. This can also be seen as a form of negative feedback.

It has been demonstrated that *oddball SGD* can speed up learning by a large factor [1] with respect to the generalisation error on the test set. However, it is not known exactly what effect *oddball SGD* has on the distribution of error across the training set. In this article, we illustrate the capacity of this learning algorithm to enforce uniformity of learning across a training set.

We train two identical instances of an associative deep neural network [see 2] to synthesise the frames of a video. This video case study is relevant because video frames are not usually uniformly distributed in their abstract feature space. Using the video frames, each DNN is independently trained with either traditional non-batch SGD or with *oddball SGD*. In order to robustly enforce uniformity of learning via *oddball SGD* [1], we raise the error magnitudes (across the training set) to a large power prior to normalised application as selection probability during *oddball SGD*. We then characterise the evolving distribution of training error across the training set for both independent (but identical) models. Our results demonstrate that *oddball SGD* may be used to strongly enforce uniform learning across the training set.

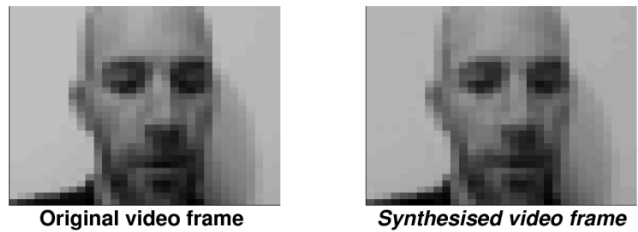

**Fig. 1. Example video frame.** We took the 32x32 pixel images (frames) and unpacked them into a vector of length 1024 to form the output at the last layer of the synthesiser DNN. Each video frame was then assigned a unique class at the input layer. On the left is plotted the original video frame and on the right is plotted a video frame synthesised using a network trained using *oddball SGD*

## II. METHOD

We consider a video sequence featuring 1000 frames, at 32x32 pixels per frame. Each pixel consisted of a grayscale intensity, normalised to the range [0,1]. We assign to each frame a unique class, giving 1000 classes. This is a form of arbitrary associative memory [2]. We then trained a DNN to synthesise each video frame (image) from the respective (unique) class. Thus, we used the DNN as a deep [2] video encoder/decoder.

The input layer to the DNN was a vector of length 1000. For each training example (frame) the respective class was set to 1 and the remainder set to 0. For the output layer we unpacked the images of 32x32 pixels into vectors of length 1024. An example video frame (and corresponding synthetic recreation) is given in Fig. 1. Pixel intensities were normalized to the range [0,1]. We built a fully connected network of size 1000x100x1024 units with sigmoid activation functions and sigmoid output layer. In terms of parameters, the original video is of dimension [1000x1024 = 1,024,000] and the synthesiser DNN features ~200,000 weights, representing a compression (dimension reduction) factor of around 5x with respect to the original video.

*Oddball SGD.* Each training iteration of *oddball SGD* began with a feed-forward pass over the 1000-element training set. Absolute prediction error (the absolute difference between the prediction of the model and the training data for the output layer) was then computed, for each training element, in the output layer with respect to the training data. Then, for each element of the training set, the sum of the absolute error (across the 1024-unit output layer) was computed and placed in a vector (length 1000) corresponding to the training examples. This vector represents the state of *novelty* of each training element [1].

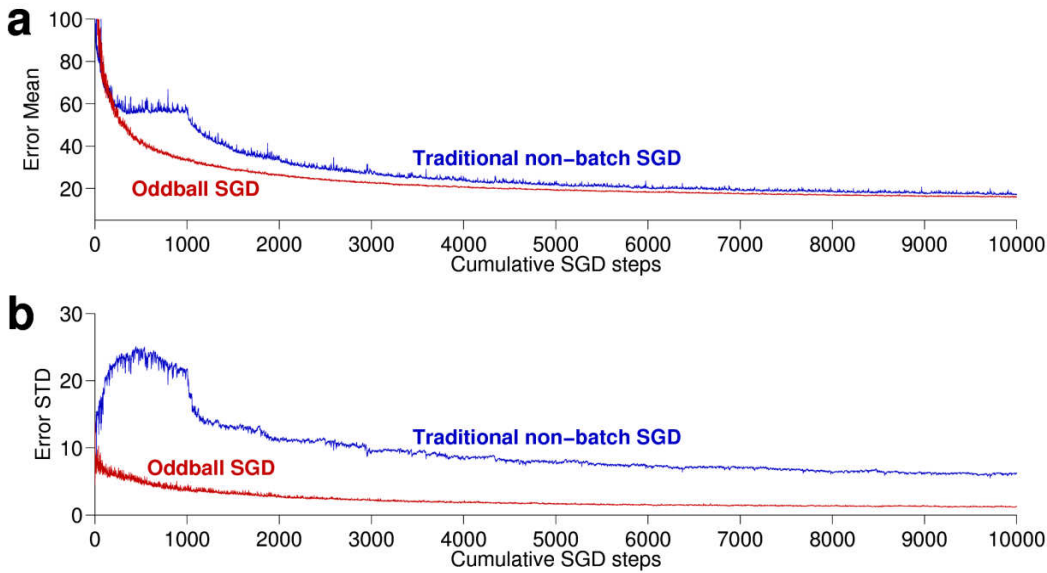

**Fig. 2. Uniform Learning via *oddball SGD*.** Summary statistics for synthesis error across the training set (1000 video frames). For each video frame, the absolute error is computed at the output layer (with respect to the original video) and summed. This gives an overall error measure for each video frame. The distribution (across the 1000 frames) is then summarised with a mean and STD. The mean is consistent with typical measures of training error. The STD captures the uniformity of error across the training set. **a** plots mean error as a function of cumulative SGD steps (1000 steps = a full sweep for traditional non-batch SGD). **b** plots error STD as a function of cumulative SGD steps (1000 steps = a full sweep for traditional non-batch SGD).

*Novelty-driven selection statistics.* In order to specifically enforce *uniformity of learning*, the *novelty* vector was raised to the power of 100. The *novelty* vector was then normalised so that it summed to 1 (i.e., it could be interpreted in terms of instantaneous probabilities). The resulting normalised *selection probability vector* was then used to assign instantaneous selection probabilities to each training element (so that selection probability was proportional to the novelty).

During each iterative step of *oddball SGD*, an element of the training set was randomly selected according to the selection probabilities. Traditional non-batch SGD was also used to train a separate model. In this case, each full-sweep iteration of training featured a random ordering of the 1000 training examples.

Each separate instance of the model (traditional non-batch SGD and *oddball SGD*) was trained for 10 full-sweep iterations (10,000 cumulative update steps of SGD respectively). Momentum was not used. At each step, the error was computed (over the 1000 video frames) and the mean and standard deviation (STD) computed. This allows us to characterise both the overall learning (mean) and the uniformity of learning (STD). For the *oddball SGD* model, SGD steps are counted cumulatively (a full sweep of SGD is 1000 steps) and are compared like for like (on a step-by-step basis). For reliable comparison, both instances of the model were trained from the exact same random starting weights. A learning rate (SGD step size) of 1 was used for all training. Dropout [4] and dither [5-7] were not used (for reasons that are beyond the scope of this article).

III. RESULTS

Our DNNs are tasked with learning a deep encoding of the video frames such that they are able to synthesise each frame on demand with as little error as possible. The difficulty is that many of the video frames are very similar, whilst the most salient frames (e.g., featuring movement) are very unusual. Hence, by the law of averages, the significance of the error in these 'oddball' frames is small. However, in terms of perception (or information), these frames are perhaps the most important in the video. Furthermore, non-uniform error (over time, in the video case) is perhaps more likely to be perceived. Hence, uniformity of error is critical for this application.

We quantify DNN performance in terms of video synthesis error (the sum of the absolute error in pixel intensity in the synthesised image, with respect to the original training image). Taking this error measure across the training set, using the mean we are able to capture the dynamics of learning in our two respective regimes. Using the STD, we are able to capture the uniformity of learning for comparison.

Fig. 2 plots the mean and STD, of summed-absolute-error, across the training examples (video frames), for the two models trained with traditional non-batch SGD and *oddball SGD* (with raised-to-the-power-of-100 *novelty vector*) respectively. Fig. 2a plots the mean summed-absolute-error across the video frames. The *oddball SGD* mean error function drops more quickly than that of the model trained with the traditional non-batch SGD method.

Fig. 2b plots the respective STD functions. As anticipated, the STD falls rapidly for the *oddball SGD* model, demonstrating the enforcement of uniformity of learning. However, the STD function actually rises, initially, for the model trained with traditional non-batch SGD and does not recover to the level of the *oddball SGD* model. Overall, the *oddball SGD* model shows a much smaller STD of error across the training set and this advantage persists throughout (and beyond 100 full sweep training iterations – data not shown).

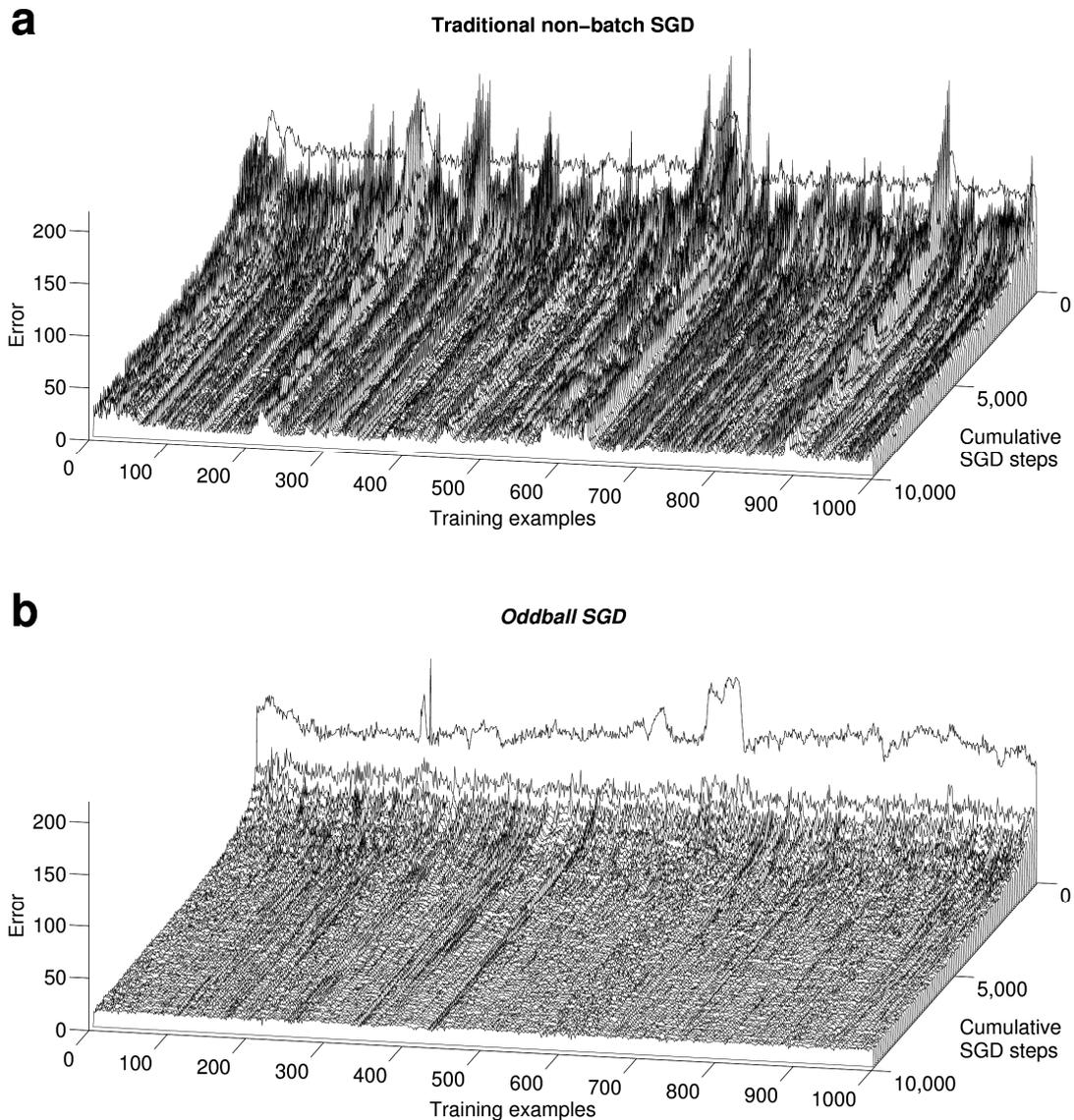

**Fig. 3. Uniform Learning via *oddball SGD: Waterfall error distribution plots*.** These plots show the sum-of-absolute error (y-axis) as distributed over the training examples (in sequential order of frames, x-axis) as a function of cumulative SGD steps (z-axis). **a** shows the evolution of the error distribution for the model trained with traditional non-batch SGD. **b** plots the same for the model trained with *oddball SGD*. NB: These error functions are not raised to any power, hence they do not linearly correspond to the raised novelty vector which drove the oddball SGD algorithm.

Fig. 3 provides 'waterfall' plots showing the actual error distributions for the respective models as they evolve over time (cumulative SGD steps). Fig. 3a shows the evolution of the distribution of error over the training examples (video frames) for the model trained with traditional non-batch SGD. There are clear non-uniformities which persist throughout the 10 full-sweep iterations (10,000 cumulative SGD steps). Generally, these error peaks in the distribution correspond to groups of frames in the video featuring movements. Thus, these distributions capture novelty and show that this novelty persists over time in a relative sense.

Fig. 3b plots the same evolution for the model trained using *oddball SGD* (with the *novelty vector* raised to the power of 100 before normalisation to create the *selection probability vector*). The error distribution is markedly uniform by comparison to that of the model trained using traditional non-batch SGD. The enforcement of uniformity is clear and robust – any peaks in the error distribution are suppressed via the negative feedback of *oddball SGD*.

## IV. Discussion and Conclusion

In this article, we have demonstrated that *oddball SGD* [1] may be used to enforce uniformity of learning across a training set. We have also noted how this might be useful for deep encoding of video or other data of non-uniform nature. In principle, the power of 100 (to which the *novelty vector* was raised) is arbitrary and acts as a tunable parameter that provides control over the uniformity of learning that is enforced.

## Acknowledgment

AJRS did this work on the weekends and was supported by his wife and children.